\pdfoutput=1
\RequirePackage{fix-cm}
\RequirePackage{etex}
\documentclass[11pt]{article}

\usepackage{authblk} 
\usepackage{eacl2023}

\usepackage{times}
\usepackage{latexsym}
\usepackage{inconsolata}

\usepackage{graphicx}
\usepackage{natbib}
\usepackage{hyperref}
\PassOptionsToPackage{hyphens}{url}

\usepackage[htt]{hyphenat}
\usepackage{url}
\usepackage[T1]{fontenc}
\usepackage{anyfontsize}
\usepackage[latin1]{inputenc}
\usepackage{microtype}
\usepackage{booktabs}

\newcommand{\bftab}{\fontseries{b}\selectfont}
\usepackage{xurl}

\usepackage{textcomp}
\usepackage{mfirstuc}

\usepackage{fontawesome}

\usepackage{array}
\usepackage{makecell}
\usepackage{multirow}
\newcolumntype{P}[1]{>{\centering\arraybackslash}p{#1}}
\newcolumntype{R}[1]{>{\raggedleft\arraybackslash}p{#1}}

\usepackage{siunitx}
\newcommand{\stddev}[1]{{\hspace{1.5pt}\raisebox{0.5pt}{\scriptsize \num{#1}}}}

\newcommand{\libraryNumQueryStrategies}{14}
\newcommand{\libraryNumStoppingCriteria}{5}

\usepackage{xspace}
\newcommand{\libraryNameText}{small-text\xspace}
\newcommand{\libraryNameTextCap}{Small-Text\xspace}
\newcommand{\libraryName}{\texttt{\libraryNameText}\xspace}

\newcommand{\githubUrl}{\url{https://github.com/webis-de/small-text}\xspace}

\newcommand{\transformersName}{\texttt{transformers}\xspace}
\newcommand{\pytorchName}{\texttt{PyTorch}\xspace}

\newcommand{\sklearnName}{\texttt{scikit-learn}\xspace}
\newcommand{\jclalName}{\texttt{JCLAL}\xspace}
\newcommand{\libactName}{\texttt{libact}\xspace}
\newcommand{\baalName}{\texttt{BaaL}\xspace}
\newcommand{\modalName}{\texttt{modAL}\xspace}
\newcommand{\lrtcName}{\texttt{lrtc}\xspace}
\newcommand{\lrtcLongName}{\texttt{low-resource-text-classification-framework}\xspace}
\newcommand{\alipyName}{\texttt{ALiPy}\xspace}
\newcommand{\scikitActivemlName}{\texttt{scikit-activeml}\xspace}

\newcommand{\altoolboxName}{\texttt{ALToolbox}\xspace}

\usepackage{xcolor} 

\graphicspath{{figures/}}
\DeclareGraphicsExtensions{.pdf,.ai,.jpg,.png}
\setkeys{Gin}{pagebox=artbox}

\newcommand{\Ni}{(i)~}
\newcommand{\Nii}{(ii)~}
\newcommand{\Niii}{(iii)~}
\newcommand{\Niv}{(iv)~}
\newcommand{\Nv}{(v)~}

\RequirePackage{type1cm}

\begin{document}

\title{\libraryNameTextCap:\kern5pt Active Learning for Text Classification in Python}

\date{}

\setlength{\affilsep}{0.7em}
\renewcommand\Authands{, }

\author[1]{\textbf{Christopher Schr{\"o}der}}
\author[1,2]{\textbf{Lydia M{\"u}ller}}
\author[1]{\textbf{Andreas Niekler}}
\author[1,3]{\textbf{Martin Potthast}}

\affil[1]{Leipzig University}
\affil[2]{Institute for Applied Informatics (InfAI), Leipzig}
\affil[3]{ScaDS.AI}

\maketitle

\begin{abstract}
We introduce \libraryName, an easy-to-use active learning library, which offers pool-based active learning for single- and multi-label text classification in Python. It features numerous pre-implemented state-of-the-art query strategies, including some that leverage the GPU. Standardized interfaces allow the combination of a variety of classifiers, query strategies, and stopping criteria, facilitating a quick mix and match, and enabling a rapid and convenient development of both active learning experiments and applications. With the objective of making various classifiers and query strategies accessible for active learning, \libraryName integrates several well-known machine learning libraries, namely \sklearnName, \pytorchName, and Hugging Face \transformersName. The latter integrations are optionally installable extensions, so GPUs can be used but are not required. Using this new library, we investigate the performance of the recently published SetFit training paradigm, which we compare to vanilla transformer fine-tuning, finding that it matches the latter in classification accuracy while outperforming it in area under the curve. The library is available under the MIT License at \githubUrl, in version~1.3.0 at the time of writing.
\end{abstract}
 
\section{Introduction}

\begin{figure}[t]
\centerline{\includegraphics[width=0.46\textwidth]{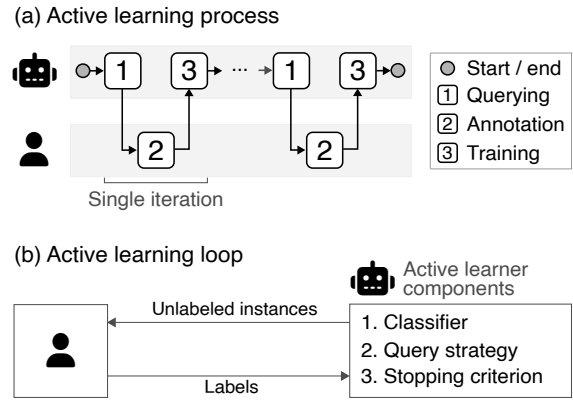}}
\caption{Illustrations of (a)~the active learning process, and (b)~the active learning setup with the components of the active learner.}
\label{fig:active-learning-setup}
\end{figure}

Text classification, like most modern machine learning applications, requires large amounts of training data to achieve state-of-the-art effectiveness. However, in many real-world use cases, labeled data does not exist and is expensive to obtain, especially when domain expertise is required. {\em Active Learning} \cite{lewis:1994} solves this problem by repeatedly selecting unlabeled data instances that are deemed informative according to a so-called {\em query strategy}, and then having them labeled by an expert (see Figure~\ref{fig:active-learning-setup}a). A new model is then trained on all previously labeled data, and this process is repeated until a specified stopping criterion is met. Active learning aims to minimize the amount of labeled data required while maximizing the effectiveness (increase per iteration) of the model, e.g., in terms of classification accuracy.
 
An active learning setup, as shown in Figure~\ref{fig:active-learning-setup}b, generally consists of up to three components on the system side: a classifier, a query strategy, and an optional stopping criterion. Meanwhile, many approaches for each of these components have been proposed and studied. Determining appropriate combinations of these approaches is only possible experimentally, and efficient implementations are often nontrivial. In addition, the components often depend on each other, for example, when a query strategy relies on parts specific to certain model classes, such as gradients \citep{ash:2020} or embeddings \citep{margatina:2021}. The more such non-trivial combinations are used together, the more the reproduction effort increases, making a modular library essential.

\begin{figure*}[t]
\centerline{\includegraphics[width=0.85\linewidth]{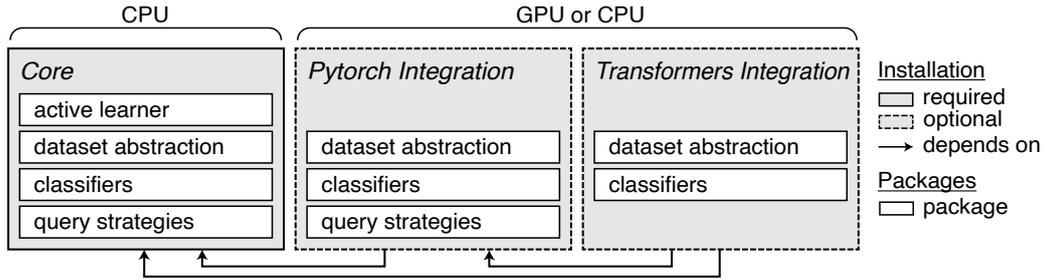}}
\caption{Module architecture of \libraryName. The core module can optionally be extended with a \pytorchName and \transformersName integration, which enable to use GPU-based models and state-of-the-art transformer-based text classifiers of the Hugging Face \transformersName library, respectively. The dependencies between the module's packages have been omitted.}
\label{fig:overview}
\end{figure*}

An obvious solution to the above problems is the use of open source libraries, which, among other benefits, accelerate research and facilitate technology transfer between researchers as well as into practice \cite{sonnenburg:2007}. While solutions for active learning in general already exist, few address text classification, which requires features specific to natural language processing, such as word embeddings \cite{mikolov:2013} or language models \cite{devlin:2019}. To fill this gap, we introduce \libraryName, an active learning library that provides tried and tested components for both experiments and applications.

\section{Overview of Small-Text}

The main goal of \libraryName is to offer state-of-the-art active learning for text classification in a convenient and robust way for both researchers and practitioners. For this purpose, we implemented a modular pool-based active learning mechanism, illustrated in Figure~\ref{fig:overview}, which exposes interfaces for classifiers, query strategies, and stopping criteria. The core of \libraryName integrates \sklearnName \cite{pedregosa:2012}, enabling direct use of its classifiers. Overall, the library provides thirteen~query strategies, including some that are only usable on text data, five~stopping criteria, and two integrations of well-known machine learning libraries, namely \pytorchName \cite{paszke:2019} and \transformersName \cite{wolf:2020}. The integrations ease the use of CUDA-based GPU computing and transformer models, respectively. The modular architecture renders both integrations completely optional, resulting in a slim core that can also be used in a CPU-only scenario without unnecessary dependencies. Given the ability to combine a considerable variety of classifiers and query strategies, we can easily build a vast number of combinations of active learning setups. 

The library provides relevant text classification baselines such as SVM \cite{joachims:1998} and KimCNN \cite{kim:2014}, and many more can be used through \sklearnName. Recent transformer models such as BERT \cite{devlin:2019} are available through the \transformersName integration. This integration also includes a wrapper that enables the use of the recently published SetFit training paradigm \cite{tunstall:2022}, which uses contrastive learning to fine-tune SBERT embeddings \cite{reimers:2019} in a sample efficient manner.

As the query strategy, which selects the instances to be labeled, is the most salient component of an active learning setup, the range of alternative query strategies provided covers four paradigms at the time of writing:
\Ni
confidence-based strategies: least confidence \cite{lewis:1994,culotta:2005}, prediction entropy \cite{roy:2001}, breaking ties~\cite{luo:2005}, BALD~\cite{houlsby:2011}, CVIRS~\cite{reyes:2018}, and contrastive active learning \cite{margatina:2021};
\Nii
embedding-based strategies: BADGE \cite{ash:2020}, BERT k-means~\cite{yuan:2020}, discriminative active learning \cite{gissin:2019}, and SEALS~\cite{coleman:2022};
\Niii
gradient-based strategies: expected gradient length~(EGL; \citealp{settles:2007}), EGL-word~\cite{zhang:2017}, and EGL-sm~\cite{zhang:2017}; and
\Niv
coreset strategies: greedy coreset~\cite{sener:2018} and lightweight coreset~\cite{bachem:2018}. Since there is an abundance of query strategies, this list will likely never be exhaustive---also because strategies from other domains, such as computer vision, are not always applicable to the text domain, e.g., when using the geometry of images \cite{konyushkova:2015}, and thus will be disregarded here. 

Furthermore, \libraryName includes a considerable amount of different stopping criteria: \Ni~stabilizing predictions \cite{bloodgood:2009}, \Niv~overall-uncertainty \cite{zhu:2008}, \Niii~classification-change \cite{zhu:2008},  \Nii predicted change of F-measure \cite{bloodgood:2019}, and \Nv~a~criterion that stops after a fixed number of iterations. Stopping criteria are often neglected in active learning although they exert a strong influence on labeling efficiency.

The library is available via the python packaging index and can be installed with just a single command: \texttt{pip install small-text}. Similarly, the integrations can be enabled using the extra requirements argument of Python's \texttt{setuptools}, e.g., the \transformersName integration is installed using \texttt{pip install small-text[transformers]}. The robustness of the implementation rests on extensive unit and integration tests. Detailed examples, an API documentation, and common usage patterns are available in the online documentation.%
\footnote{\url{https://small-text.readthedocs.io}}

\section{Library versus Annotation Tool}

We designed \libraryName for two types of settings: \Ni experiments, which usually consist of either automated active learning evaluations or short-lived setups with one or more human annotators, and \Nii real-world applications, in which the final model is subsequently applied on unlabeled or unseen data. Both cases benefit from a library which offers a wide range of well-tested functionality.

To clarify on the distinction between a library and an annotation tool, \libraryName is a library, by which we mean a reusable set of functions and classes that can be used and combined within more complex programs. In contrast, annotation tools provide a graphical user interface and focus on the interaction between the user and the system. Obviously, \libraryName is still intended to be used by annotation tools but remains a standalone library. In this way it can be used \Ni in combination with an annotation tool, \Nii within an experiment setting, or \Niii as part of a backend application, e.g. a web API. As a library it remains compatible to all of these use cases. This flexibility is supported by the library's modular architecture which is also in concordance with software engineering best practices, where high cohesion and low coupling \cite{myers:1975} are known to contribute towards highly reusable software \cite{mueller:1993,tonella:2001}. 
As a result, \libraryName should be compatible with most annotations tools that are extensible and support text classification.

\section{Code Example}

In this section we show a code example to perform active learning with transformers models.

\paragraph{Dataset} First, we create (for the sake of a simple example) a synthetic two-class spam dataset of $100$ instances. The data is given by a list of texts and a list of integer labels. To define the tokenization strategy, we provide a \transformersName tokenizer. From these individual parts we construct a \texttt{TransformersDataset} object which is a dataset abstraction that can be used by the interfaces in \libraryName. This yields a binary text classification dataset containing 50 examples of the positive class (spam) and the negative class (ham) each:

\vspace*{0.01em}
\noindent\includegraphics[viewport=0 6.8cm 7.7cm 15cm, clip]{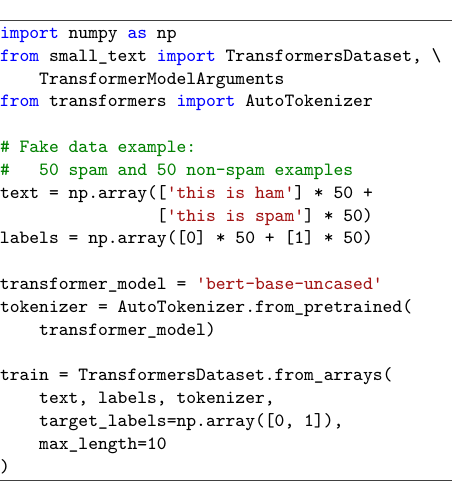}

\paragraph{Active Learning Configuration} Next, we configure the classifier and query strategy. Although the active learner, query strategies, and stopping criteria components are dataset- and classifier-agnostic, classifier and dataset have to match (i.e. \texttt{TransformerBasedClassification} must be used with \texttt{TransformersDataset}) owing to the different underlying data structures:

\begin{figure}[h!]\includegraphics[viewport=0 10.4cm 7.7cm 14.68cm, clip]{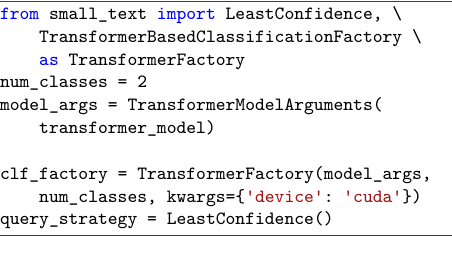}
\end{figure}

\noindent Since the active learner may need to instantiate a new classifier before the training step, a factory \cite{gamma:1995} is responsible for creating new classifiers. Finally, we set the query strategy to least confidence \cite{culotta:2005}.

\paragraph{Initialization} There is a chicken-and-egg problem for active learning   because most query strategies rely on the model, and a model in turn is trained on labeled instances which are selected by the query strategy. This problem can be solved by either providing an initial model (e.g. through manual labeling), or by using cold start approaches \cite{yuan:2020}. In this example we simulate a user-provided initialization by looking up the respective true labels and providing an initial model:

\noindent\includegraphics[viewport=0 8.9cm 7.7cm 15cm, clip]{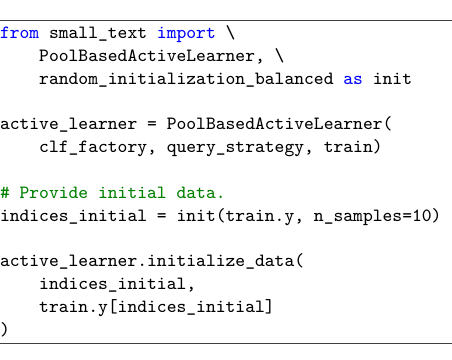}

\noindent To provide an initial model in the experimental scenario (where true labels are accessible), \libraryName provides sampling methods, from which we use the balanced sampling to obtain a subset whose class distribution is balanced (or close thereto). In a real-world application, initialization would be accomplished through a starting set of labels supplied by the user. Alternatively, a cold start classifier or query strategy can be used instead.

\paragraph{Active Learning Loop} After the previous code examples prepared the setting by loading a dataset, configuring the active learning setup, and providing an initial model, the following code block shows the actual active learning loop. In this example, we perform five queries during each of which ten instances are queried. During a query step the query strategy samples instances to be labeled. Subsequently, new labels for each instance are provided and passed to the \texttt{update} method, and then a new model is trained. In this example, it is a simulated response relying on true labels, but in a real-world application this part is the user's response.

\noindent\includegraphics[viewport=0 5.8cm 7.7cm 15cm, clip]{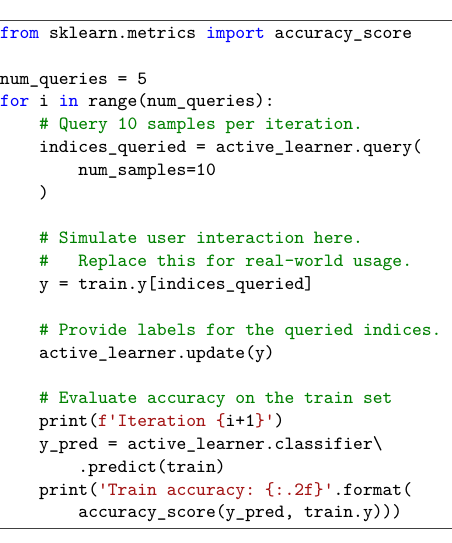}

\noindent In summary, we built a full active learning setup in only very few lines of code. The actual active learning loop consists of just the previous code block and changing hyperparameters, e.g., using a different query strategy, is as easy as adapting the \texttt{query\_strategy} variable.

\section{Comparison to Previous Software}

\begin{table*}[!t]
\small
\centering
\renewcommand{\tabcolsep}{8.4pt}
\begin{tabular}{@{}lrrcccllcc@{}}
\toprule
  \bfseries Name               &            \multicolumn{3}{c}{\bfseries Active Learning}             &                                                     \multicolumn{6}{c@{}}{\bfseries Code}                                                      \\
 \cmidrule(l@{\tabcolsep}r@{\tabcolsep}){2-4} \cmidrule(l@{\tabcolsep}){5-10} 
                               &                         QS &                          SC &   Text    &    GPU    &   Unit    & Language & License      &  Last  &                                       Reposi-                                       \\
                               &                            &                             &   Focus   &  support  &   Tests   &          &              & Update &                                        tory                                         \\
\midrule
  \texttt{JCLAL}$^1$           &                         18 &                           2 & \faRemove & \faRemove & \faRemove & Java     & GPL          &  2017  &                 \href{https://github.com/ogreyesp/JCLAL}{\faGithub}                 \\
  \texttt{libact}$^2$          &                         19 &                           - & \faRemove & \faRemove & \faCheck  & Python   & BSD-2-Clause &  2021  &                \href{https://github.com/ntucllab/libact}{\faGithub}                 \\
  \texttt{modAL}$^3$           &                         21 &                           - & \faRemove & \faCheck  & \faCheck  & Python   & MIT          &  2022  &               \href{https://github.com/modAL-python/modAL}{\faGithub}               \\
  \texttt{ALiPy}$^4$           &                         22 &                           4 & \faRemove & \faRemove & \faCheck  & Python   & BSD-3-Clause &  2022  &                 \href{https://github.com/NUAA-AL/ALiPy}{\faGithub}                  \\
  \texttt{BaaL}$^5$            &                          9 &                           - & \faRemove & \faCheck  & \faCheck  & Python   & Apache 2.0   &  2023  &                 \href{https://github.com/baal-org/baal}{\faGithub}                  \\
  \texttt{lrtc}$^6$            &                          7 &                           - & \faCheck  & \faCheck  & \faRemove & Python   & Apache 2.0   &  2021  & \href{https://github.com/IBM/low-resource-text-classification-framework}{\faGithub} \\
  \texttt{scikit-activeml}$^7$ &                         29 &                           - & \faRemove & \faCheck  & \faCheck  & Python   & BSD-3-Clause &  2023  &        \href{https://github.com/scikit-activeml/scikit-activeml}{\faGithub}         \\
  \texttt{ALToolbox}$^8$       &                         19 &                           - & \faCheck  & \faCheck  & \faCheck  & Python   & MIT          &  2023  &           \href{https://github.com/AIRI-Institute/al_toolbox}{\faGithub}            \\
\midrule
  \texttt{\libraryName}        & \libraryNumQueryStrategies & \libraryNumStoppingCriteria & \faCheck  & \faCheck  & \faCheck  & Python   & MIT          &  2023  &              \href{https://github.com/webis-de/small-text}{\faGithub}               \\
\bottomrule
\end{tabular}
\caption{Comparison between \libraryName and relevant previous active learning libraries. We abbreviated the number of query strategies by ``QS'', the number of stopping criteria by ``SC'', and the \texttt{low-resource-text-classification} framework by \texttt{lrtc}. All information except ``Publication Year'' and ``Code Repository'' has been extracted from the linked GitHub repository of the respective library on February~24th, 2023. Random baselines were not counted towards the number of query strategies. Publications: $^1$\citet{reyes:2016}, $^2$\citet{yang:2017}, $^3$\citet{danka:2018}, $^4$\citet{tang:2019}, $^5$\citet{atighehchian:2020}, $^6$\citet{ein-dor:2020}, $^7$\citet{kottke:2021}, $^8$\citet{tsvigun:2022}.}
\label{table-related-software}
\end{table*}

Unsurprisingly, after decades of research and development on active learning, numerous other libraries are available that focus on active learning as well. In the following we present a selection of the most relevant open-source projects for which either a related publication is available or a larger user base exists: \jclalName \cite{reyes:2016} is a generic framework for active learning which is implemented in Java and can be used either through XML configurations or directly from the code. It offers an experimental setting which includes 18 query strategies. The aim of \libactName \cite{yang:2017} is to provide active learning for real-world applications. Among 19 other strategies, it includes a well-known meta-learning strategy \cite{hsu:2015}. \baalName \cite{atighehchian:2020} provides bayesian active learning including methods to obtain uncertainty estimates. The \modalName library \cite{danka:2018} offers single- and multi-label active learning, provides 21 query strategies, also builds on \sklearnName by default, and offers instructions how to include GPU-based models using \texttt{Keras} and \pytorchName. \alipyName \cite{tang:2019} provides an active learning framework targeted at the experimental active learning setting. Apart from providing 22~query strategies, it supports alternative active learning settings, e.g., active learning with noisy annotators. The \lrtcLongName (\lrtcName; \cite{ein-dor:2020}) is an experimentation framework for the low resource scenario and supports which can be easily extended. It also focuses on text classification and has a number of built-in models, datasets, and query strategies to perform active learning experiments. Another recent library is \scikitActivemlName which offers general active learning built around \sklearnName. It comes with 29 query strategies but provides no stopping criteria. GPU-based functionality can be used via \texttt{skorch},\footnote{We also evaluated the use of \texttt{skorch} but transformer models were not supported at that time.} a \pytorchName wrapper, which is a ready-to-use adapter as opposed to our implemented classifier structures but is on the other hand restricted to the \sklearnName interfaces. \altoolboxName~\cite{tsvigun:2022} is an active learning framework that provides an annotation interface and a benchmarking mechanism to develop new query strategies. While it has some overlap with \libraryName, it is not a library, but also focuses on text data, namely on text classification and sequence tagging.

In Table~\ref{table-related-software}, we compare \libraryName to the previously mentioned libraries, and compare them based on several criteria related to active learning or to the respective code base: While all libraries provide a selection of query strategies, not all libraries offer stopping criteria, which are crucial to reducing the total annotation effort and thus directly influence the efficiency of the active learning process \cite{vlachos:2008,laws:2008,olsson:2009}. We can also see a difference in the number of provided query strategies. While a higher number of query strategies is certainly not a disadvantage, it is more important to provide the most relevant strategies (either due to recency, domain-specificity, strong general performance, or because it is a baseline). Based on these criteria, \libraryName provides numerous recent strategies such as BADGE~\cite{ash:2020}, BERT K-Means~\cite{yuan:2020}, and contrastive active learning~\cite{margatina:2021}, as well as the gradient-based strategies by \citet{zhang:2017}, where the latter are unique to active learning for text classification. Selecting a subset of query strategies is especially important since active learning experiments are computationally expensive \cite{margatina:2021,schroeder:2022}, and therefore not every strategy can be tested in the context of an experiment or application. Finally, only \libraryName{}, \lrtcName, and \altoolboxName focus on text, and only about half of the libraries offer access to GPU-based deep learning, which has become indispensable for text classification due to the recent advances and ubiquity of transformer-based models \cite{vaswani:2017,devlin:2019}. 

The distinguishing characteristic of \libraryName is the focus on text classification, paired with a multitude of interchangeable components. It offers the most comprehensive set of features (as shown in Table~\ref{table-related-software}) and through the integrations these components can be mixed and matched to easily build numerous different active learning setups, with or without leveraging the GPU. Finally, it allows to use concepts from natural language processing (such as transformer models) and provides query strategies unique to text classification. 

\section{Experiment}

\begin{table}[t!]%
\centering
\fontsize{9pt}{10pt}\selectfont%
\renewcommand{\tabcolsep}{4pt}%
\begin{tabular}[t]{@{}l@{\hspace{5pt}}ccr@{\hspace{3pt}}r@{}}
\toprule
\bfseries Dataset Name {\tiny (ID)} & \bfseries Type & \bfseries Classes & \bfseries Training & \bfseries Test \\
\midrule
AG's News$^1$ {\scriptsize (AGN)}       & N & 4 &  120,000 & \textsuperscript{$\star$}7,600 \\
Customer Reviews$^2$ {\scriptsize (CR)} & S   & 2 &    3,397 &                         378 \\
Movie Reviews$^3$ {\scriptsize (MR)}    & S   & 2 &    9,596 &                       1,066 \\
Subjectivity$^4$ {\scriptsize (SUBJ)}   & S   & 2 &    9,000 &                       1,000 \\
TREC-6$^5$ {\scriptsize (TREC-6)}       & Q   & 6 &    5,500 & \textsuperscript{$\star$}500 \\
\bottomrule
\end{tabular}
\caption{Key characteristics about the examined datasets: $^1$\citet{zhang:2015}, $^2$\citet{hu:2004}, $^3$\citet{pang:2005}, $^4$\citet{pang:2004}, $^5$\citet{li:2002}. The dataset type was abbreviated by N (News), S (Sentiment), Q (Questions). $\star$:~Predefined test sets were available and adopted.}
\label{table-datasets}
\end{table}

We perform an active learning experiment comparing an SBERT model trained with the recent sentence transformers fine-tuning paradigm (SetFit; \cite{tunstall:2022}) over a BERT model trained with standard fine-tuning. 
SetFit is a contrastive learning approach that trains on pairs of (dis)similar instances. Given a fixed amount of differently labeled instances, the number of possible pairs is considerably higher than the size of the original set, making this approach highly sample efficient~\cite{chuang:20,henaff:20} and therefore interesting for active learning. 

\paragraph{Setup} We reproduce the setup of our previous work \cite{schroeder:2022} and evaluate on the datasets shown in Table~\ref{table-datasets} with an extended set of query strategies. Starting with a pool-based active learning setup with 25 initial samples, we perform 20~queries during each of which 25 instances are queried and labeled. Since SetFit has only been evaluated for single-label classification~\cite{tunstall:2022}, we focus on single-label classification as well. The goal is to compare the following two models: \Ni~BERT (\texttt{bert-large-uncased}; \cite{devlin:2019}) with 336M parameters and \Nii~SBERT (\texttt{paraphrase-mpnet-base-v2}; \cite{reimers:2019})  with 110M parameters. The first model is trained via vanilla fine-tuning and the second using SetFit. For the sake of brevity, we refer to the first as ``BERT'' and to the second as ``SetFit''. To compare their performance during active learning, we provide an extensive benchmark over multiple computationally inexpensive uncertainty-based query strategies, which were selected due to  encouraging results in our previous work. Moreover, we include BALD, BADGE, and greedy coreset---all of which are computationally more expensive, but have been increasingly used in recent work  \cite{ein-dor:2020,yu:2002}.

\begin{table}[!t]%
\centering
\fontsize{9pt}{10pt}\selectfont%
\renewcommand{\tabcolsep}{8pt}%
\begin{tabular}[t]{@{}lccccc@{}}
\toprule 
  {\bfseries Model} & \bfseries Strategy & \multicolumn{2}{c}{\bfseries Rank} & \multicolumn{2}{c@{}}{\bfseries Result} \\
\cmidrule(l@{\tabcolsep}r@{\tabcolsep}){3-4}\cmidrule(l@{\tabcolsep}){5-6}
                    &                    & Acc.         & AUC                 & Acc.          & AUC                     \\
\midrule
  BERT              & PE                 & 2.20         & 2.80                & 0.917         & 0.858                   \\
                    & BT                 & \bftab{1.40} & \bftab{1.60}        & \bftab{0.919} & \bftab{0.868}           \\
                    & LC                 & 3.80         & 3.20                & 0.916         & 0.863                   \\
                    & CA                 & 4.20         & 5.00                & 0.915         & 0.857                   \\
                    & BA                 & 3.00         & 5.20                & 0.917         & 0.855                   \\
                    & BD                 & 6.20         & 4.60                & 0.909         & 0.862                   \\
                    & CS                 & 6.60         & 7.60                & 0.910         & 0.843                   \\
                    & RS                 & 7.80         & 5.40                & 0.901         & 0.856                   \\
\midrule
  SetFit            & PE                 & 2.80         & 3.20                & 0.927         & 0.906                   \\
                    & BT                 & 2.80         & \bftab{1.60}        & 0.926         & \bftab{0.912}           \\
                    & LC                 & \bftab{2.20} & 2.60                & \bftab{0.927} & 0.908                   \\
                    & CA                 & 4.80         & 3.80                & 0.924         & 0.907                   \\
                    & BA                 & 5.20         & 6.20                & 0.923         & 0.902                   \\
                    & BD                 & 6.60         & 5.60                & 0.915         & 0.904                   \\
                    & CS                 & 2.80         & 4.40                & 0.927         & 0.909                   \\
                    & RS                 & 6.60         & 6.80                & 0.907         & 0.899                   \\
\bottomrule
\end{tabular}
\caption{The ``Rank'' columns show the mean rank when ordered by mean accuracy (Acc.) and by area under curve (AUC). The ``Result'' columns show the mean accuracy and AUC. All values used in this table refer to state after the final iteration. Query strategies are abbreviated as follows: prediction entropy (PE), breaking ties (BT), least confidence (LC), contrastive active learning (CA), BALD (BA), BADGE (BD), greedy coreset (CS), and random sampling (RS).}
\label{table-results-summary}
\end{table}

\paragraph{Results}

In Table~\ref{table-results-summary}, the results show the summarized classification performance in terms of \Ni~final accuracy after the last iteration, and \Nii~area under curve (AUC). We also compare strategies by ranking them from 1~(best) to 8~(worst) per model and dataset by accuracy and AUC.  First, we can also confirm for SetFit the earlier finding that uncertainty-based strategies perform strong for BERT \cite{schroeder:2022}. Second, SetFit configurations result in between 0.06 and 1.7 percentage points higher mean accuracy, and also in betwen 4.2 and 6.6 higher AUC when averaged over model and query strategy. Interestingly, the greedy coreset strategy (CS) is remarkably more successful for the SetFit runs compared to the BERT runs.

\noindent Detailed results per configuration can be found in the appendix, where it can be seen that SetFit reaches higher accuracy scores in most configurations, and better AUC scores in all configurations.

\begin{figure}[!t] 
\includegraphics[width=0.482\textwidth]{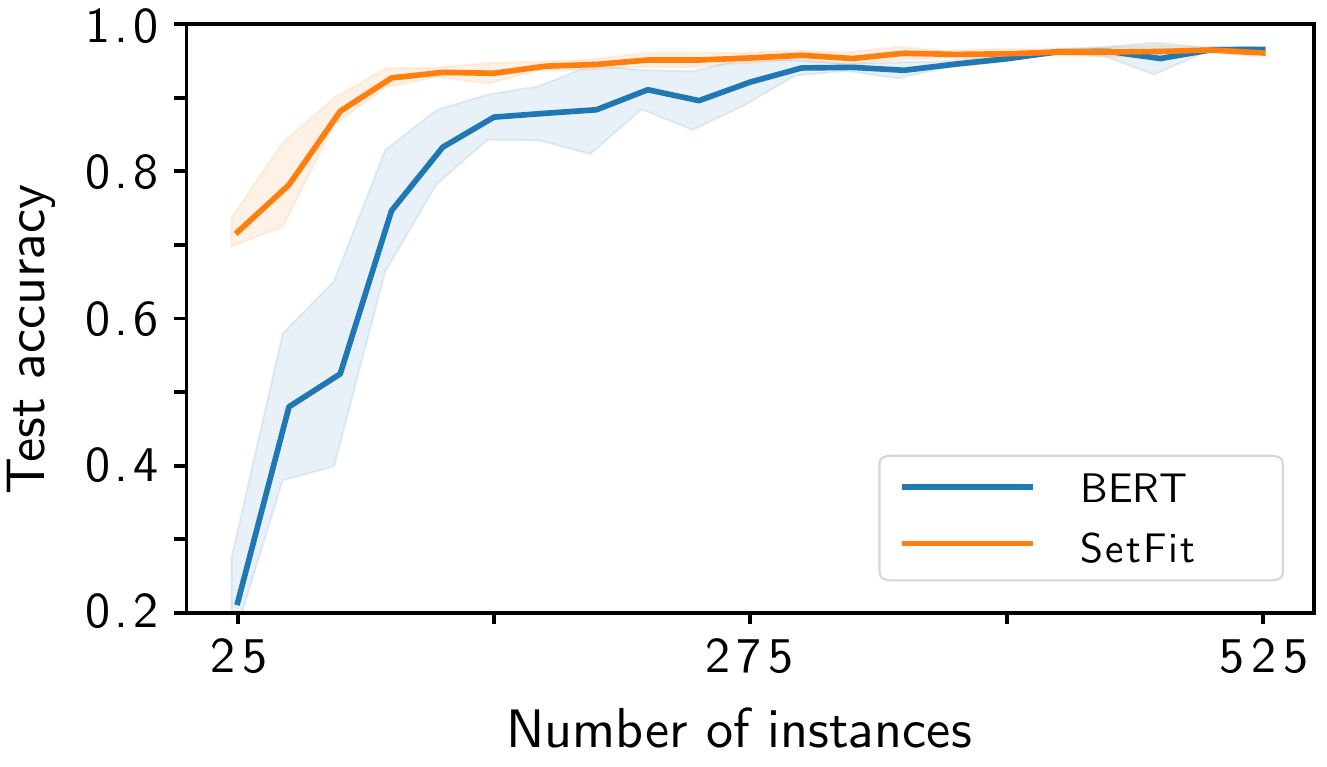}
\caption{An exemplary learning curve showing the difference in test accuracy for breaking ties strategy on the TREC dataset, comparing BERT and SetFit. The tubes represent the standard deviation across five runs.}
\label{fig:learning-curve}
\end{figure}

\paragraph{Discussion} When trained with the new SetFit paradigm, models having only a third of the parameters compared to the large BERT model achieve results that are not only competitive, but slightly better regarding final accuracy and considerably better in terms of AUC. Since the final accuracy values are often within one percentage point or less to each other, it is obvious that the improvement in AUC stems from improvements in earlier queries, i.e. steeper learning curves. We suspect that this is at least partly owed to sample efficiency from SetFit's training that uses pairs of instances. Moreover, this has the additional benefit of reducing instability of transformer models \cite{mosbach:2021} as can be exemplarily seen in Figure~\ref{fig:learning-curve}. This increasingly occurs when the training set is small \cite{mosbach:2021}, which is likely alleviated with the additional instance pairs. On the other hand, training cost increase linearly with the number of pairs per instance. In the low-data regime, however, this is a manageable additional cost that is worth the benefits.

\section{Library Adoption}
\label{library-adoption}

As recent publications have already adopted \libraryName, we present four examples which have already successfully utilized it for their experiments.

\paragraph{Abusive Language Detection} \citet{kirk:2022} investigated the detection of abusive language using transformer-based active learning on six datasets of which two exhibited a balanced and four an imbalanced class distribution. They evaluated a pool-based binary active learning setup, and their main finding is that, when using active learning, a model for abusive language detection can be efficiently trained using only a fraction of the data.

\paragraph{Classification of Citizens' Contributions} In order to support the automated classification of  German texts from online citizen participation processes, \citet{romberg:2022} used active learning to classify texts collected by three cities into eight different topics. They evaluated this real-world dataset both as a single- and multi-label active learning setup, finding that active learning can considerably reduce the annotation efforts.
 
\paragraph{Softmax Confidence Estimates} \citet{gonsior:2022} examined several alternatives to the softmax function to obtain better confidence estimates for active learning. Their setup extended \libraryNameText to incorporate additional softmax alternatives and found that confidence-based methods mostly selected outliers. As a remedy to this they proposed and evaluated uncertainty clipping.

\paragraph{Revisiting Uncertainty-Based Strategies} In a previous publication, we reevaluated traditional uncertainty-based query strategies with recent transformer models \cite{schroeder:2022}. We found that uncertainty-based methods can still be highly effective and that the breaking ties strategy is a drop-in replacement for prediction entropy.\\

\noindent Not only have all of these works successfully applied \libraryNameText to a variety of different problems, but each work is also accompanied by a GitHub repository containing the experiment code, which is the outcome we had hoped for. We expect that \libraryName will continue to gain adoption within the active learning and text classification communities, so that future experiments will increasingly rely on it by both reusing existing components and by creating their own extensions, thereby supporting the field through open reproducible research.

\section{Conclusion}

We introduced \libraryName, a modular Python library, which offers state-of-the-art active learning for text classification. It integrates \sklearnName, \pytorchName, and \transformersName, and provides robust components that can be mixed and matched to quickly apply active learning in both experiments and applications, thereby making active learning easily accessible to the Python ecosystem. 

\newpage
\section*{Limitations}

Although a library can, among other things, lower the barrier of entry, save time, and speed up research, this can only be leveraged with basic knowledge of the Python programming language. All included algorithmic components are subject to their own limitations, e.g., the greedy coreset strategy quickly becomes computationally expensive as the amount labeled data increases. Moreover, some components have hyperparameters which require an understanding of the algorithm to achieve the best classification performance. In the end, we provide a powerful set of tools which still has to be properly used to achieve the best results.

As \libraryName covers numerous text classification models, query strategies, and stopping criteria, some limitations from natural language processing, text classification and active learning apply as well. For example, all included classification models rely on tokenization, which is inherently more difficult for languages which have no clear word boundaries such as Chinese, Japanese, Korean, or Thai.

\section*{Ethics Statement}

In this paper, we presented \libraryName, a library which can---like any other software---be used for good or bad. It can be used to bootstrap classification models in scenarios where no labeled data is available. This could be used for good, e.g. for spam detection, hatespeech detection, or targeted news filtering, but also for bad, e.g., for creating models that detect certain topics that are to be censored in authoritarian regimes. While such systems already exist and are of sophisticated quality, \libraryName is unlikely to change anything at this point. On the contrary, being open-source software, these methods can now be used by a larger audience, which contributes towards the democratization of classification algorithms.
\section*{Acknowledgments}

We thank the anonymous reviewers for their constructive advice and the early adopters of the library for their invaluable feedback. 

This research was partially funded by the Development Bank of Saxony (SAB) under project numbers 100335729 and 100400221. 
Computations were done (in part) using resources of the Leipzig University Computing Centre.

\vskip 0.2in

\bibliography{eacl23-small-text-library-lit}
\bibliographystyle{acl_natbib}

\appendix

\begin{table*}[!ht]%
\centering
\fontsize{9pt}{10pt}\selectfont%
\renewcommand{\tabcolsep}{5.6pt}%
\begin{tabular}{@{}ll@{\hspace{10pt}}llllllll@{}}
\toprule
\textbf{Dataset} & \textbf{Model} & \multicolumn{8}{c}{\textbf{Query Strategy}}\\
\cmidrule{3-10} && \multicolumn{1}{c}{PE}& \multicolumn{1}{c}{BT}& \multicolumn{1}{c}{LC}& \multicolumn{1}{c}{CA}& \multicolumn{1}{c}{BA}& \multicolumn{1}{c}{BD}& \multicolumn{1}{c}{CS}& \multicolumn{1}{c}{RS}\\
\midrule
\multirow{2}{*}{AGN} & BERT   & 0.898\stddev{0.003}  & 0.901\stddev{0.004}  & 0.900\stddev{0.001}  & 0.889\stddev{0.010}  & 0.889\stddev{0.008}  & 0.894\stddev{0.003}  & 0.881\stddev{0.006}  & 0.886\stddev{0.004}\\
    & SetFit   & 0.900\stddev{0.002}  & 0.902\stddev{0.004}  & \bftab{0.902}\stddev{0.002}  & 0.892\stddev{0.006}  & 0.887\stddev{0.010}  & 0.896\stddev{0.003}  & 0.896\stddev{0.003}  & 0.877\stddev{0.005}\\
\midrule
\multirow{2}{*}{CR} & BERT   & 0.920\stddev{0.009}  & 0.920\stddev{0.009}  & 0.916\stddev{0.006}  & 0.917\stddev{0.010}  & 0.919\stddev{0.010}  & 0.911\stddev{0.010}  & 0.915\stddev{0.012}  & 0.902\stddev{0.014}\\
    & SetFit   & 0.937\stddev{0.014}  & 0.937\stddev{0.014}  & 0.937\stddev{0.014}  & 0.938\stddev{0.009}  & 0.934\stddev{0.004}  & 0.913\stddev{0.011}  & \bftab{0.939}\stddev{0.011}  & 0.912\stddev{0.010}\\
\midrule
\multirow{2}{*}{MR} & BERT   & 0.850\stddev{0.005}  & 0.850\stddev{0.005}  & 0.846\stddev{0.008}  & 0.844\stddev{0.008}  & 0.859\stddev{0.003}  & 0.835\stddev{0.017}  & 0.843\stddev{0.006}  & 0.831\stddev{0.020}\\
    & SetFit   & 0.871\stddev{0.009}  & 0.871\stddev{0.009}  & 0.871\stddev{0.009}  & 0.869\stddev{0.004}  & 0.867\stddev{0.005}  & 0.864\stddev{0.008}  & 0.870\stddev{0.008}  & \bftab{0.871}\stddev{0.003}\\
\midrule
\multirow{2}{*}{SUBJ} & BERT   & 0.959\stddev{0.005}  & 0.959\stddev{0.005}  & 0.958\stddev{0.003}  & 0.958\stddev{0.008}  & 0.959\stddev{0.003}  & 0.948\stddev{0.006}  & 0.957\stddev{0.004}  & 0.937\stddev{0.006}\\
    & SetFit   & 0.962\stddev{0.004}  & 0.962\stddev{0.004}  & 0.962\stddev{0.004}  & 0.960\stddev{0.002}  & \bftab{0.966}\stddev{0.002}  & 0.942\stddev{0.002}  & 0.963\stddev{0.003}  & 0.932\stddev{0.005}\\
\midrule
\multirow{2}{*}{TREC-6} & BERT   & 0.960\stddev{0.002}  & 0.966\stddev{0.003}  & 0.960\stddev{0.008}  & 0.965\stddev{0.006}  & 0.958\stddev{0.007}  & 0.958\stddev{0.009}  & 0.952\stddev{0.015}  & 0.947\stddev{0.009}\\
    & SetFit   & 0.966\stddev{0.005}  & 0.961\stddev{0.005}  & 0.966\stddev{0.005}  & 0.963\stddev{0.008}  & 0.961\stddev{0.005}  & 0.958\stddev{0.005}  & \bftab{0.967}\stddev{0.004}  & 0.946\stddev{0.009}\\
\bottomrule
\end{tabular}
\caption{%
Final accuracy per dataset, model, and query strategy. We report the mean and standard deviation over five runs. The best result per dataset is printed in bold. Query strategies are abbreviated as follows: prediction entropy (PE), breaking ties (BT), least confidence (LC), contrastive active learning (CA), BALD (BA), BADGE (BD), greedy coreset (CS), and random sampling (RS). The best result per dataset is printed in bold.}
\label{table-classification-results-acc}
\end{table*}

\section*{Supplementary Material}

\section{Technical Environment}
 
All experiments were conducted within a Python 3.8 environment. The system had CUDA 11.1 installed and was equipped with an {NVIDIA GeForce RTX 2080 Ti} (11GB VRAM). 

\section{Experiments}

Each experiment configuration represents a combination of model, dataset and query strategy, and has been run for five times. 

\subsection{Datasets}

We used datasets that are well-known benchmarks in text classification and active learning. All datasets are accessible to the Python ecosystem via Python libraries that provide fast access to those datasets. We obtained CR and SUBJ using \href{https://nlp.gluon.ai}{gluonnlp}, and  AGN, MR, and TREC using \href{https://github.com/huggingface/datasets}{huggingface datasets}.

\begin{table*}[!ht]%
\centering
\fontsize{9pt}{10pt}\selectfont%
\renewcommand{\tabcolsep}{5.6pt}%
\begin{tabular}{@{}ll@{\hspace{10pt}}llllllll@{}}
\toprule
\textbf{Dataset} & \textbf{Model} & \multicolumn{8}{c}{\textbf{Query Strategy}}\\
\cmidrule{3-10} && \multicolumn{1}{c}{PE}& \multicolumn{1}{c}{BT}& \multicolumn{1}{c}{LC}& \multicolumn{1}{c}{CA}& \multicolumn{1}{c}{BA}& \multicolumn{1}{c}{BD}& \multicolumn{1}{c}{CS}& \multicolumn{1}{c}{RS}\\
\midrule
\multirow{2}{*}{AGN} & BERT   & 0.827\stddev{0.009}  & 0.839\stddev{0.014}  & 0.836\stddev{0.009}  & 0.821\stddev{0.015}  & 0.819\stddev{0.012}  & 0.840\stddev{0.003}  & 0.804\stddev{0.012}  & 0.825\stddev{0.011}\\
    & SetFit   & 0.881\stddev{0.002}  & \bftab{0.889}\stddev{0.003}  & 0.885\stddev{0.005}  & 0.879\stddev{0.004}  & 0.869\stddev{0.006}  & 0.881\stddev{0.002}  & 0.881\stddev{0.003}  & 0.867\stddev{0.004}\\
\midrule
\multirow{2}{*}{CR} & BERT   & 0.885\stddev{0.007}  & 0.885\stddev{0.007}  & 0.881\stddev{0.007}  & 0.881\stddev{0.011}  & 0.882\stddev{0.006}  & 0.876\stddev{0.005}  & 0.874\stddev{0.011}  & 0.877\stddev{0.011}\\
    & SetFit   & 0.925\stddev{0.001}  & 0.925\stddev{0.001}  & 0.925\stddev{0.001}  & 0.927\stddev{0.003}  & 0.924\stddev{0.005}  & 0.910\stddev{0.005}  & \bftab{0.930}\stddev{0.002}  & 0.908\stddev{0.008}\\
\midrule
\multirow{2}{*}{MR} & BERT   & 0.819\stddev{0.010}  & 0.819\stddev{0.010}  & 0.820\stddev{0.007}  & 0.813\stddev{0.009}  & 0.817\stddev{0.013}  & 0.808\stddev{0.011}  & 0.804\stddev{0.010}  & 0.813\stddev{0.004}\\
    & SetFit   & 0.859\stddev{0.004}  & 0.859\stddev{0.004}  & 0.859\stddev{0.004}  & \bftab{0.859}\stddev{0.003}  & 0.858\stddev{0.004}  & 0.855\stddev{0.002}  & 0.858\stddev{0.004}  & 0.857\stddev{0.002}\\
\midrule
\multirow{2}{*}{SUBJ} & BERT   & 0.944\stddev{0.008}  & 0.944\stddev{0.008}  & 0.943\stddev{0.007}  & 0.940\stddev{0.009}  & 0.939\stddev{0.009}  & 0.929\stddev{0.005}  & 0.934\stddev{0.006}  & 0.924\stddev{0.007}\\
    & SetFit   & \bftab{0.953}\stddev{0.002}  & \bftab{0.953}\stddev{0.002}  & \bftab{0.953}\stddev{0.002}  & 0.952\stddev{0.003}  & 0.950\stddev{0.002}  & 0.940\stddev{0.003}  & 0.949\stddev{0.001}  & 0.935\stddev{0.002}\\
\midrule
\multirow{2}{*}{TREC-6} & BERT   & 0.818\stddev{0.033}  & 0.855\stddev{0.023}  & 0.837\stddev{0.034}  & 0.829\stddev{0.030}  & 0.816\stddev{0.029}  & 0.856\stddev{0.024}  & 0.799\stddev{0.037}  & 0.843\stddev{0.008}\\
    & SetFit   & 0.910\stddev{0.008}  & \bftab{0.934}\stddev{0.005}  & 0.919\stddev{0.008}  & 0.917\stddev{0.013}  & 0.907\stddev{0.017}  & 0.934\stddev{0.010}  & 0.927\stddev{0.008}  & 0.927\stddev{0.004}\\
\bottomrule
\end{tabular}
\caption{%
Final area under curve (AUC) per dataset, model, and query strategy. We report the mean and standard deviation over five runs. The best result per dataset is printed in bold. Query strategies are abbreviated as follows: prediction entropy (PE), breaking ties (BT), least confidence (LC), contrastive active learning (CA), BALD (BA), BADGE (BD), greedy coreset (CS), and random sampling (RS). The best result per dataset is printed in bold.}
\label{table-classification-results-auc}
\end{table*}

\subsection{Pre-Trained Models}

In the experiments, we fine-tuned \Ni~a large BERT model (\href{https://huggingface.co/bert-large-uncased}{bert-large-uncased}) and \Nii~an SBERT paraphrase-mpnet-base model (\href{https://huggingface.co/sentence-transformers/paraphrase-mpnet-base-v2}{sentence-transformers/paraphrase-mpnet-base-v2}). Both are available via the \href{https://huggingface.co/models}{huggingface model repository}.

\subsection{Hyperparameters}

\paragraph{Maximum Sequence Length}

We set the maximum sequence length to the minimum multiple of ten, so that 95\% of the given dataset's sentences contain at most that many tokens.

\paragraph{Transformer Models}

For BERT, we adopt the hyperparameters from \citet{schroeder:2022}. For SetFit, we use the same learning rate and optimizer parameters but we train for only one epoch.

\section{Evaluation}

In Table~\ref{table-classification-results-acc} and Table~\ref{table-classification-results-auc} we report final accuracy and AUC scores including standard deviations, measured after the last iteration. Note that results obtained through PE, BT, and LC are equivalent for binary datasets.

\subsection{Evaluation Metrics}

Active learning was evaluated using standard metrics, namely accuracy und area under the learning curve. For both metrics, the respective scikit-learn implementation was used.

\newpage
\section{Library Adoption}

As mentioned in Section~\ref{library-adoption}, the experiment code of previous works documents how \libraryName was used and can be found at the following locations:

\begin{itemize}
\setlength{\itemsep}{1pt}
\item
Abusive Language Detection:\\ \url{https://github.com/HannahKirk/ActiveTransformers-for-AbusiveLanguage}
\item
Classification of Citizens' Contributions:\\ \url{https://github.com/juliaromberg/egov-2022}
\item
Softmax Confidence Estimates:\\
\url{https://github.com/jgonsior/btw-softmax-clipping}
\item
Revisiting Uncertainty-Based Strategies:\\
\url{https://github.com/webis-de/ACL-22}
\end{itemize}

\end{document}